\documentclass{article}
\usepackage{amsmath,graphicx,subfigure,amssymb,flushend,spconf}
\usepackage{algorithm,algorithmic,color,colortbl}
\usepackage{booktabs}
\usepackage{colortbl}

\newcommand{\CL}{\cellcolor[gray]{0.8}}

\usepackage{fancyheadings}
\title{Gabor Barcodes for Medical Image Retrieval}

\name{Mina Nouredanesh$^1$, Hamid R. Tizhoosh$^{2*}$, Ershad Banijamali$^3$}  
\address{$^1$ Department of Mechanical and Mechatronics Engineering, University of Waterloo, ON, Canada\\
$^2$ KIMIA Lab, University of Waterloo, Canada, \emph{tizhoosh@uwaterloo.ca} [$^*$corresponding author]\\
$^3$ Cheriton School of Computer Science, University of Waterloo, ON, Canada}

\flushend
\begin{document}
\maketitle

\begin{abstract}
In recent years, advances in medical imaging have led to the emergence of massive databases, containing images from a diverse range of modalities. This has significantly heightened the need for automated annotation of the images on one side, and fast and memory-efficient content-based image retrieval systems on the other side. Binary descriptors have recently gained more attention as a potential vehicle to achieve these goals. One of the recently introduced binary descriptors for tagging of medical images are Radon barcodes (RBCs) that are driven from Radon transform via local thresholding.  Gabor transform is also a  powerful transform to extract texture-based information. Gabor features have exhibited robustness against rotation, scale, and also photometric disturbances, such as illumination changes and image noise in many applications. This paper introduces Gabor Barcodes (GBCs), as a novel framework for the image annotation. To find the most discriminative GBC for a given query image, the effects of employing Gabor filters with different parameters, i.e., different sets of scales and orientations, are investigated, resulting in different barcode lengths and retrieval performances. The proposed method has been evaluated on the IRMA dataset with 193 classes comprising of 12,677 x-ray images for indexing, and 1,733 x-rays images for testing. A total error score as low as $351$ ($\approx 80\%$ accuracy for the first hit) was achieved.
\end{abstract}


\section{Introduction}
Recent developments in medical imaging and widespread use of picture archiving and communication systems (PACS), have provided unique opportunities for image-based diagnosis, inter-patient comparisons, and searching for the images, sharing characteristics similar to the region of interest (ROI), i.e., tumours. According to \cite{1}, a single average size radiology department represents tens of terabytes of data every year. The massive databases have heightened the need for efficient data storage (for long-term archiving) and retrieval methods. In recent years, many studies have focused on the development of fast query search systems, using binary features, e.g., binary hashing. Recently, the concept of ``barcode annotation'' has been proposed as a method for content-based medical image retrieval \cite{2,2a}. This is a deviation of the established notation as we generally understand \emph{annotations} to be textual metadata. Barcodes can be added to the DICOM (Digital Image and Communication in Medicine) files to provide supplementary information, in order to increase the accuracy of image retrieval. The possibility of applying Radon barcodes (RBCs) to encode regions of interest (ROI Barcodes) also offers a new and exciting opportunity to collect the diary fingerprint of lesions and tissue types \cite{2}. The other strength of the proposed Radon barcodes is their efficiency in terms of retrieval speed (due to the possibility of using Hamming distance and/or hashing) and lower requirements toward storage space. In the present paper, we aim to further develop the idea of ``barcodes'' from a different point of view and to generate texture-based barcodes for the task of content-based medical image retrieval.

\section{Background}

\begin{figure}[htbp]
\begin{center}
\includegraphics[width=0.70\columnwidth]{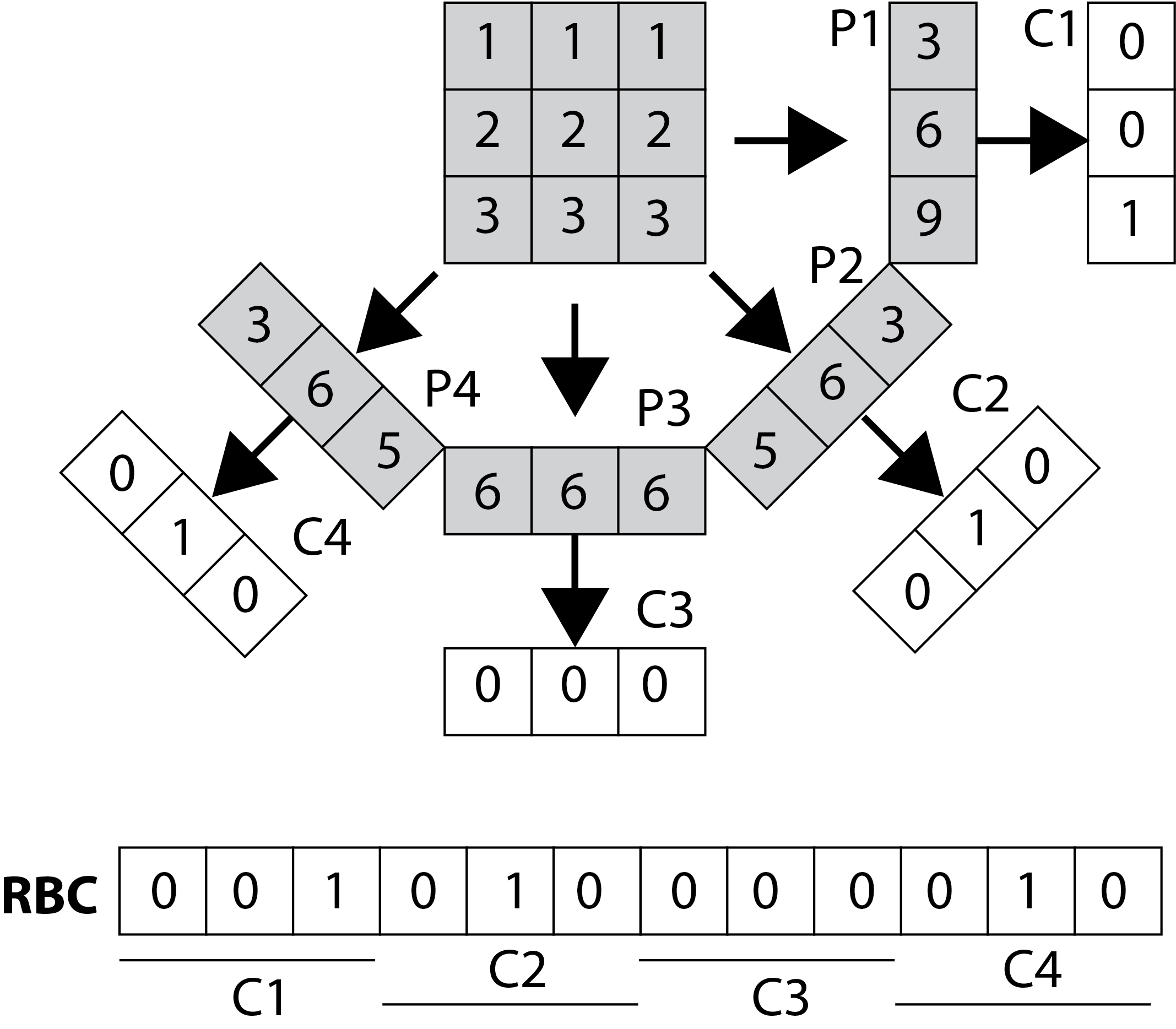}
\caption{Radon Barcode (RBC) \cite{2} -- All Radon projections (P1,P2,P3,P4) are thresholded to generate code fragments C1,C2,C3,C4. The concatenation of all code fragments delivers the barcode \textbf{RBC}. }
\label{fig:RBC}
\end{center}
\end{figure}

Tizhoosh introduced the notion of using Radon transform to generate a content-based barcode \cite{2}. Looking at an image $I$ as a function $f(x,y)$, one can project $f(x,y)$ along a number of projection angles. The projection is basically the sum (integral) of $f(x,y)$ values along lines constituted by each angle $\theta$. The projection creates a new image $R(\rho,\theta)$ with $\rho = x \cos \theta + y \sin \theta$. Hence, using the Dirac delta function $\delta(\cdot)$ the Radon transform can be written as 
\begin{equation}
R(\rho,\theta) = \int\limits_{-\infty}^{+\infty} \int\limits_{-\infty}^{+\infty} f(x,y) \delta(\rho-x\cos \theta-y\sin\theta) dx dy.
\end{equation}
If we threshold all projections (lines) for individual angles based on a ``local'' threshold for that angle, then we can assemble a barcode of all thresholded projections as depicted in Figure \ref{fig:RBC}. A simple way for thresholding the projections is to calculate a typical value via median operator applied on all non-zero values of each projection \footnote{Matlab code available online: http://tizhoosh.uwaterloo.ca/}. 

On the other hand, texture analysis has been an active research and numerous algorithms has been proposed based on different models, e.g., grey-level co-occurrence (GLC) matrices and Markov random field (MRF) model \cite{3,4}. In recent works, wavelets have become very popular due to their capacity to provide multi-resolution analysis of the images. Among different kinds of the wavelet transform, the Gabor transform has some interesting mathematical and biological properties (resembling the characteristics of human visual cortical cells) and has been widely used to extract texture features from images for either segmentation tasks \cite{7, 8}, object detection and biometric identification \cite{13, 14}, and image retrieval \cite{9,10,11,12}. 

In \cite{5}, the authors have compared the effectiveness of various texture analysis and classification methods such as dyadic wavelet, wavelet frame, Gabor wavelet, and steerable pyramids, and have observed that the Gabor-based methods outperform the others on textured images. Moreover, the performance of Gabor wavelet features for texture analysis is investigated, and compared with other features, i.e., tree-structured wavelet transform, showing that Gabor features provide the best pattern retrieval accuracy compared to other multiresolution texture features on the Brodatz texture database \cite{6}.

The most important property of Gabor features is their robustness against rotation, scale, and translation. Furthermore, they are robust against photometric disturbances, such as illumination changes and noise. These properties are mainly due to the fact that the parameters of Gabor filters enable us to establish invariance in this regard \cite{15}. 

\section{Gabor transform} 
In the spatial domain, a two-dimensional Gabor filter is a Gaussian function, modulated by a exponential or complex sinusoidal plane wave, defined as
\begin{equation}
G(x,y) = \frac{f^2}{\pi\gamma\eta}\exp{\left( -\frac{x'^2+\gamma y'^2}{2\sigma^2}\right)}\exp{(j2\pi f  x'+\phi)}
\end{equation}
where $x'=x\cos\theta+y\sin\theta$, $y'=-x\sin\theta+y\cos\theta$, $f$ is the frequency of the sinusoid (modulation frequency), $\theta$ represents the orientation of the normal to the parallel stripes of a Gabor function,  $\phi $ is the phase offset, $\sigma$ is the standard deviation of the Gaussian envelope, and $\gamma$ is the spatial aspect ratio which specifies the ellipticity of the support of the Gabor function \cite{14}. Given an image $I(x,y)$, the response of Gabor filter is the convolution of Gabor window with image $I$ given by 
\begin{equation}
\psi_{u,v}(x,y) = \sum_s\sum_t I(x-s,y-t)*G_{u,v}(s,t)
\end{equation}
where $s$ and $t$ are the window/mask size of the Gabor filter, $u$ is the number of scales and $v$ is the number of orientations that are used in the Gabor filter bank (GFB$(u,v,s,t)$). The $\psi_{u,v}(x,y)$ forms complex valued function including real and imaginary parts. In this study, in order to obtain Gabor features, the magnitudes of the $\psi_{u,v}$ values ($\psi_{ABS-u,v}$) are calculated. There have been several studies in the literature reporting the optimal values for the parameters of the Gabor filter bank (i.e., spatial frequencies and number of orientations) in such a way that it can mimic human visual system as much as possible \cite{16}. 

\section{Gabor Barcodes}
\begin{figure}[htb]
\begin{center}
\vspace{0.01in}
\includegraphics[width=0.7\columnwidth]{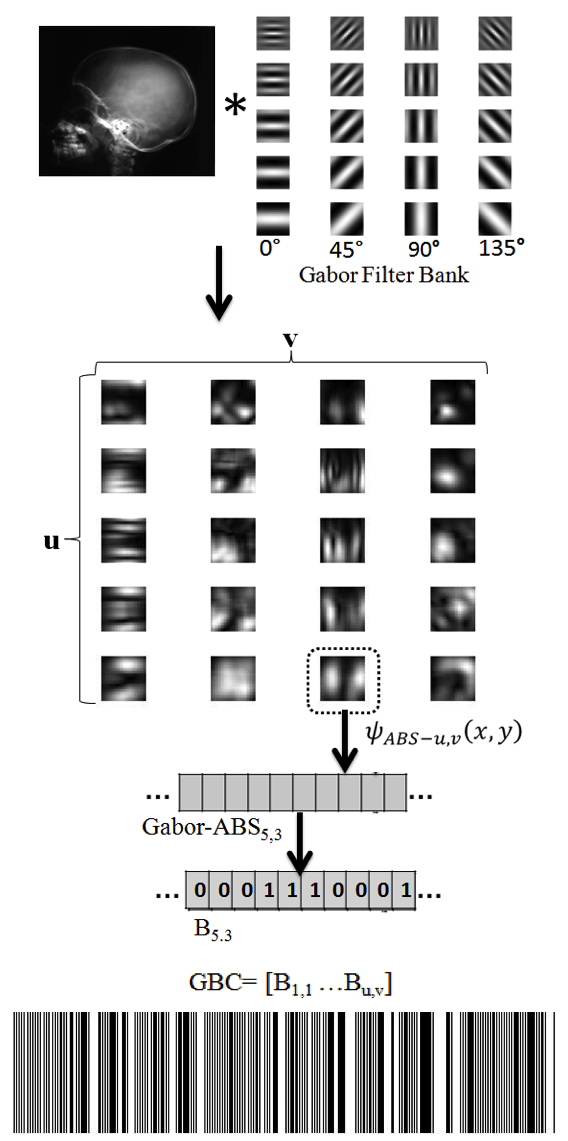}
\caption{Generation of Gabor barcodes by binarizing and appending the Gabor feature vectors, Median values are selected as the binarization threshold }
\label{fig:fig1}
\end{center}
\end{figure}

In order to obtain the Gabor feature vectors with the same-length, images should be resized into $R_N\times C_N$ images, i.e., $R_N=C_N=2^n\in\mathbb{N}^+$ (in this study  $32\times 32$ images). To generate the Gabor Barcode (GBC$_m$) for the query image $I_m$, after obtaining the magnitude of each filtered responses ($\psi_{ABS-u,v}(x,y)$), for every $u$ and $v$ in the Gabor filter bank, the 2-D matrices of $\psi_{ABS-u,v}(x,y)$, are first downsampled with the coefficient factor of 4, and transformed to the row vectors of real-value Gabor feature (Gabor-ABS$_{u,v,m}$). For each (Gabor-ABS$_{u,v,m}$ vector, the median ($T_{u,v,m}$) is calculated and employed as a threshold to binarize the corresponding feature vector and obtain $B_{u,v,m}$ (same approach to binarize Radon barcodes in \cite{2}). The final GBC$_m$ extracted for  $I_m$ is obtained by concatenating all $u\!\times\!v$ binary vectors (Fig. \ref{fig:fig1}).

 Employing a GFB with $N_g$ Gabor filters ($N_g= u\times v$), the dimension of the feature vector before downsampling is $M\times N\times N_g$ (e.g., for $N_g=40$ and $M=N=32$, the dimension of the feature vector is 40960). Since the adjacent pixels in an image are usually highly correlated, it is possible to reduce this redundancy by downsampling the feature images. The features were downsampled by a factor of $d1$ and $d2$ for the column and row, respectively, in which $d1=d2=4$ (the downsampled feature vector will have a size of $\frac{40960}{4\!\times\!4} =2560$ for forty Gabor filters). Generally, the Gabor feature vector of a $M\times N$ image is a column vector with length $(M\times N\times u\times v)/(d1\times d2)$ \cite{14}. The steps of the approach are described in Algorithm \ref{alg:alg1}.

\begin{algorithm}[htb]
\vspace{0.01in}
\caption{Generation of Gabor Barcodes (inspired by \cite{2})}
\begin{algorithmic}[1]
\label{alg:alg1}
\STATE Initialize Gabor Barcode for image $I_m$: GBC$_m \leftarrow \empty$
\STATE Initialize $R_N =C_N \leftarrow 32$
\STATE $I =$ Normalize($I , R_N , C_N$)
\STATE Apply Gabor filters with $u$ scales $v$ orientations  
\FOR{all $u$ and $v$} 
	\STATE Calculate the magnitude of  $\psi_{u,v}(x,y)$: \\ $\psi_{ABS-u,v}(x,y) = |\psi_{u,v}(x,y)|$
	\STATE Downsample each $\psi_{ABS-u,v}(x,y)$ with factor of 4 
	\STATE  Generate row feature vectors Gabor-ABS$_{u,v,m}$ 
	\STATE Typical$_{u,v,m} \leftarrow$ median(Gabor-ABS$_{u,v,m}$)
	\STATE $ B_{u,v,m} \leftarrow $ Gabor-ABS$_{u,v,m} \geq$ Typical$_{u,v,m}$
	\STATE GBC$_m \leftarrow$ append(GBC$_m$, B$_{u,v,m}$)
\ENDFOR
\end{algorithmic}
\end{algorithm}

\section{Experimental Settings}
\textbf{Data set --} The Image Retrieval in Medical Applications (IRMA) 2009 database is a collection of 14,410 radiographic images belonging to 193 different categories that have been randomly collected from daily routine work at the Department of Diagnostic Radiology of the RWTH Aachen University, from different ages, genders, view positions, and pathologies \cite{H2,H3}. For indexing, 12,677 images are available whereas  1,733 additional images can be used for testing. Each image in the dataset is annotated with an ``IRMA code'', corresponding to one of the 193 classes. The IRMA code comprises of four axes with three to four positions: 1) the technical code (T) (modality), 2) the directional code (D) (body orientations), 3) the anatomical code (A) (body region), and 4) the biological code (B) (the biological system examined). The complete IRMA code consists of 13 characters: TTTT--DDD--AAA--BBB. The IRMA database contains x-ray images that are captured at different positions and have major variations including intensity, contrast, scale, angle and so forth. Moreover, there are high overlaps between different classes. The 193 different IRMA codes are not uniformly distributed, and some codes have a considerably larger share compared to the other classes in the training set (imbalanced data distribution). These are the factors that make the IRMA database a challenging one for the CBIR task. 

\textbf{Error measurement and evaluation scheme --} We used the evaluation scheme provided by ImageCLEFmed09 to compute the difference between the IRMA codes of the testing image and the first hit retrieved by the proposed approach. The total error for all test images can then be calculates as follows \cite{18}:
\begin{equation} \label{equation:IRMA1}
E_\textrm{total} = \sum_{m=1}^{1733} \sum_{j=1}^{4} \sum_{i=1}^{l_{j}} \frac{1} {b_{l_{j},i}} \frac {1} {i} \delta (I_{l_{j},i}^{m}, \tilde{I}_{l_{j},i}^{m})
\end{equation}

Here, $m$ is an indicator to each image, $j$ is an indicator to the structure of an IRMA code, and $l_{j}$ refers to the number of characters in each structure of an IRMA code.
For example, in the IRMA code: 1121-4a0-914-700, $l_{1}=4$, $l_{2}=3$, $l_{3}=3$ and $l_{4}=3$. 
$i$ is an indicator to a character in a particular structure. 
Here, $l_{2,2}$ refers to the character ``a'' and $l_{4,1}$ refers to the character ``7''. 
$b_{l_{j},i}$ refers to the number of branches, i.e. number of possible characters, at the position $i$ in the $l_{j}^{th}$ structure in an IRMA code. 
$I^{m}$ refers to the $m^{th}$ testing image and $\tilde{I}^{m}$ refers to its top 1 retrieved image. 
$\delta (I_{l_{j},i}^{m}, \tilde{I}_{l_{j},i}^{m})$ compares a particular position in the IRMA code of the testing image and the retrieved image. 
It then outputs a value in \{0, 1\} according to the following rules.

\begin{equation} \label{equation:IRMA2}
\delta (I_{l_{j},i}^{m}, \tilde{I}_{l_{j},i}^{m})=
\begin{cases}
0, & I_{l_{j},h}^{m} = \tilde{I}_{l_{j},h}^{m} \forall h \leq i \\
1, & I_{l_{j},h}^{m} \neq \tilde{I}_{l_{j},h}^{m} \exists h \leq i
\end{cases}
\end{equation}

\section{Results}
For testing, 1,733 IRMA images were used. For each of the test images complete search was performed to find the most similar image in the Hamming space of GBCs, whereas the similarity of an input image $I_i^{\textrm{query}}$ annotated with the corresponding barcode GBC$_i^{\textrm{query}}$ is calculated based on the Hamming distance to any other image $I_j$  in the training set, with its annotated barcode GBC$_j$. The most similar case can be retrieved via
\begin{equation}
\max_{j=1,2,3,\dots,1733; j\neq i} \left( 1 - \frac{| \textrm{XOR} (\textrm{GBC}_i^{\textrm{query}},\textrm{GBC}_j)|} {\textrm{GBC}_i^{\textrm{query}}} \right)
\end{equation}

Different Gabor filters are based on different filter window size $\in \{5\times 5,11\times 11, 21\times 21, 23\times 23, 27\times 27\}$, in ($u$) scales and ($v$) orientations with $v\in\{4,8,12,16,20\}$ were employed. Based on our investigations, the Gabor filters with the window size of 23 pixels performs well on the resized images of $32\times 32$ pixels. The results are reported in Table \ref{tab:results}. For the sake of comparison, the results of CBIR reported in \cite{2}, base on the barcodes generated via local binary patterns (LBP) and local Radon binary pattern (LRBP) and Radon barcodes (RBC) are also reported.

\begin{table*}[htb]
\centering
\caption{Comparing the results of GBC with results reported in \cite{2} using two rankings based on $E_\textrm{total}$ and $\eta$.}
\label{tab:results}
\begin{tabular}{l | lll || llllllll}
Rank & Barcode      & $E_\textrm{total}$   & $L_\textrm{code}$ & Rank & Barcode      & $E_\textrm{total}$   & $L_\textrm{code}$ & $\eta$         \\ \hline 
1 & GBC8,16,23,23   & \CL 351.798  & 8192  & 1 & RBC4            & 476.62   & 512   &  \CL16.85065671 \\
2 & GBC8,12,23,23   & \CL 356.603  & 6144  & 2 & RBC8            & 478.54   & 1024  &  \CL8.39152422  \\
3 & GBC5,12,23,23   & \CL 361.739  & 3840  & 3 & GBC(5,4,23,23)  & 416.5496 & 1280  &  \CL7.71227244  \\
4 & GBC5,20,23,23   & \CL 364.1979 & 6400  & 4 & GBC(5,8,23,23)  & 364.973  & 2560  &  \CL4.401070764 \\
5 & GBC(5,8,23,23)  & \CL 364.973  & 2560  & 5 & GBC(5,8,27,27)  & 372.85   & 2560  &  \CL4.308091726 \\
6 & GBC(5,16,23,23) & \CL 365.0334 & 5120  & 6 & RBC16           & 470.57   & 2048  &  \CL4.266825339 \\
7 & GBC(5,8,27,27)  & \CL 372.85   & 2560  & 7 & GBC(5,8,11,11)  & 444      & 2560  &  \CL3.61772973  \\
8 & GBC(10,8,23,23) & \CL 374.422  & 5120  & 8 & GBC5,12,23,23   & 361.739  & 3840  &  \CL2.96027799  \\
9 & GBC(5,4,23,23)  & \CL 416.5496 & 1280  & 9 & GBC(5,16,23,23) & 365.0334 & 5120  &  \CL2.200171272 \\
10 & GBC(5,8,11,11)  & \CL 444      & 2560  & 10 & GBC(10,8,23,23) & 374.422  & 5120  &  \CL2.145002163 \\
11 & LBP             & \CL 463.81   & 7200  & 11 & RBC32           & 475.92   & 4096  &  \CL2.109430156 \\
12 & RBC16           & \CL 470.57   & 2048  & 12 & GBC8,12,23,23   & 356.603  & 6144  &  \CL1.876821003 \\
13 & RBC32           & \CL 475.92   & 4096  & 13 & GBC5,20,23,23   & 364.1979 & 6400  &  \CL1.764174917 \\
14 & RBC4            & \CL 476.62   & 512   & 14 & GBC8,16,23,23   & 351.798  & 8192  &  \CL1.42684154  \\
15 & RBC8            & \CL 478.54   & 1024  & 15 & LBP             & 463.81   & 7200  &  \CL1.231363992 \\
16 & LRBP4           & \CL 483.54   & 7200  & 16 & LRBP4           & 483.54   & 7200  &  \CL1.181120349 \\
17 & LRBP32          & \CL 501.96   & 7200  & 17 & LRBP32          & 501.96   & 7200  &  \CL1.137777778
\end{tabular}
\end{table*}

\begin{figure}[htbp]
\begin{center}
\includegraphics[width=0.7\columnwidth]{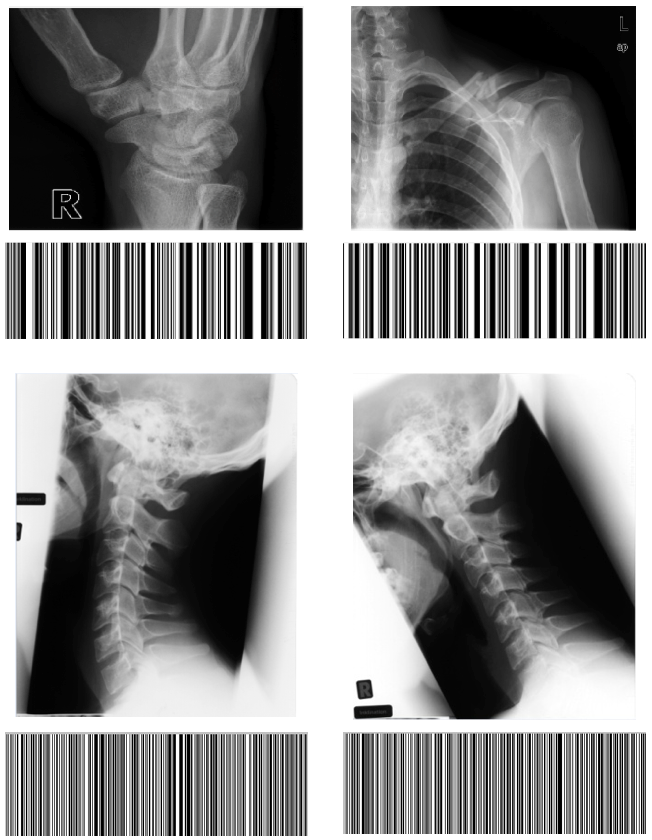}
\caption{Generated Gabor Barcodes for four images from IRMA database}
\label{fid:fig4}
\end{center}
\end{figure}

The length of the barcode is of utmost importance, playing an important in memory consumption and efficient archiving of the data, specifically in massive databases. Using the code length $L_{\textrm{code}}$, one can establish a suitability measure $\eta$ that prefers low error and short codes simultaneously:
\begin{equation}
\eta^k = \frac{\max\limits_i (E_{\textrm{total}}^i) \times \max\limits_i  (L_{\textrm{code}}^i)}{E_{\textrm{total}}^k \times L_{\textrm{code}}^k}
\end{equation}
Apparently, the larger $\eta$, the better the method, a desired quantification if the code length is  important in computation. The average time for extraction of the GBRs was around $0.150$ seconds. In contrast, average time RBC was around $0.007$ seconds (measurements on a Intel core i7 with 3.60GHz). 

\section{Conclusions}
Inspired by Radon barcodes, we introduced the notion of ``Gabor Barcodes'' in his paper. We tested their performance for content-based image retrieval using IRMA dataset. The results show that Gabor barcodes can be quite accurate for retrieving the first hit from a large archive. In our future work, we will work on ROI-based image retrieval.

\end{document}